# Validate and Enable Machine Learning in Industrial AI

Hongbo Zou*, Guangjing Chen*, Pengtao Xie*, Sean Chen*, Yongtian He*, Hochih Huang*, Zheng Nie*, Hongbao Zhang*, Tristan Bala*, Kazi Tulip*, Yuqi Wang*, Shenlin Qin*, Eric P. Xing*[†]

*Petuum Inc., Sunnyvale, US

[†]School of Computer Science, Carnegie Mellon University, Pittsburgh, US

Email: {hongbo.zou, guangjing.chen, pengtao.xie}@petuum.com

## ABSTRACT

Industrial Artificial Intelligence (Industrial AI) is an emerging concept which refers to the application of artificial intelligence to industry. Industrial AI promises more efficient future industrial control systems. However, manufacturers and solution partners need to understand how to implement and integrate an AI model into the existing industrial control system. A well-trained machine learning (ML) model provides many benefits and opportunities for industrial control optimization; however, an inferior Industrial AI design and integration limits the capability of ML models. To better understand how to develop and integrate trained ML models into the traditional industrial control system, test the deployed AI control system, and ultimately outperform traditional systems, manufacturers and their AI solution partners need to address a number of challenges. Six top challenges, which were real problems we ran into when deploying Industrial AI, are explored in the paper. The Petuum Optimum system is used as an example to showcase the challenges in making and testing AI models, and more importantly, how to address such challenges in an Industrial AI system.

## 1 Introduction

"Industrial AI" is an emerging concept which refers to the application of artificial intelligence to industry [27]. Industrial AI is now underway, transforming traditional industrial control into smart Industry 4.0 and creating new opportunities, where an AI solution is built on pattern recognition with the ability to understand those processes, interact with the environment, and intelligently adapt their behavior [22][26]. The concept of Industrial AI is becoming widely adopted in mainstream manufacturers in their discourse and practice. The recent popularity of Industrial AI is due to the increase in deployed sensors along with the automated data collection process, the powerful computation capability to perform complex tasks, faster network connection to access cloud services for data management, and computing power outsourcing. However, ML technology alone cannot be applied to industrial automation immediately if not sufficiently adapted to the industry domain. According to a Gartner survey in 2018, 75% of early AI projects underwhelmed, and 85% of AI projects failed [17]. This trend is due in large part to the understanding gap between AI system design and user expectations.

To understand how AI works with industrial control systems, we first review the framework of traditional industrial control systems without AI assistance (Figure 1). The system consists of five blocks, which include a supervisor, a controller, an actuator, a plant and measuring sensors. The supervisor is responsible for defining the desired values and instructing the connected controller. The controller detects the error signal and amplifies it to trigger the predefined control logic. The output of the controller is fed to the actuator for control logic execution. An actuator is a component of a machine that is responsible for moving and controlling that mechanism. It also acts as the input for the plant according to the control signal, so that the output signal approaches the reference input signal. The sensors, or measurement elements, are devices that convert output variables, such as displacement, into another manageable variable, such as a voltage, that can be used to compare the output with the reference input signal. This completes the feedback path of the closed-loop system [4].

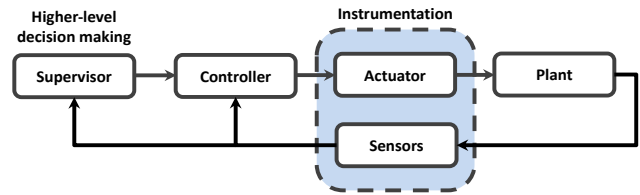

Figure 1 A typical industrial control system

Automatic control is one of the primary goals of industrial process applications. An automatic controller compares the real values of the output of a plant with the input references (the desired values), determines the deviation and produces a control signal that will reduce the deviation to zero or a small value [1]. The way in which the automatic controller produces the control signal is called control action. It moves a system from its initial condition to the desired state and, once there, maintains the desired state. AI has helped fundamentally change the scope and pace of automation. In the past, mechanical or electrical hardware components performed most control functions in technological systems. When hardware solutions were insufficient, continuous human participation in the control loop was necessary. In Industrial AI, machine learning enables predictions to take over many control functions in the supervisor and controller block. A well-trained machine learning model can provide excellent system performance under widely varying operating conditions.

Despite the strength of Industrial AI for future industrial control systems, ML models alone are not enough. Comprehensive designing and testing processes are necessary to ensure its adaptation in the industry. To better understand how to develop machine learning models, integrate them into the traditional industrial control system, and validate the deployed AI control system, manufacturers and their AI solution partners need to address all requirements and challenges. For example, a machine learning model makes suggestions based on its prediction to avoid undesired outcomes before they appear. This proactive process differs from traditional human control when people usually only react to outcomes in post-actions. Another challenge is that the ML models trained over mostly normal datasets can never guarantee to fully satisfy all of the edge cases in industrial control [11]. Therefore, the traditional control logic still needs to coexist with AI-control for extreme situations. This coexistence brings new questions, such as how to define the working scopes of AI control and traditional control logic in the industrial system, and whether this coexistence introduces conflicts in the control system. All of these questions need to be explored before Industrial AI implementation.

To help enterprises at every step of the AI adoption journey, we built Petuum Symphony, an enterprise AI platform. Petuum Symphony handles AI/ML operations, data processing, AI hardware optimization, and explainable AI allowing users to rapidly build, deploy, and maintain cutting-edge AI solutions. Operationalized on the Symphony platform, Petuum Optimum, our flagship Industrial AI product, enables manufacturing enterprises to reach unprecedented yield and energy savings by optimizing asset and process operations. By abstracting and sharing our experience in designing and implementing Petuum AI in real projects, we showcase the challenges in making and testing AI models, and more importantly, how to address these challenges in an Industrial AI system.

This paper makes the following contributions:

1. The difference of traditional industrial control and Industrial AI control is analyzed and presented;

2. Six implementation challenges, which were real problems we ran into when deploying Industrial AI, are explored and discussed in the paper;

3. Petuum Optimum, out flagship Industrial AI product, is used as an example to showcase the challenges in making and testing;

4. Introduce how to design and implement various ML and system testing approaches for thorough Industrial AI testing.

Petuum Optimum took the listed contributions and enabled two industrial AI solutions, which have been evaluated and deployed, working for two of largest cement factories in the world. The rest of this paper is organized as follows. Related work is discussed in Section 2. A tradition control system is introduced and discussed in related work. Section 3 is problems and Petuum solutions. Six top challenges are proposed and explained in this section. Also, this section introduces an Industrial AI validation system adopted by Petuum Optimum. This validation system gives us a concrete example to present our design and test experiences. Then, the six challenges are addressed and discussed, respectively, from Section 4 to Section 9. The conclusions, limitations, and future works are discussed in Section 10.

## 2 Related Work

Thanks to recent advances in machine learning and AI, Industrial AI is an emerging field with a great amount of data, market value and new industry revolution. Industrial AI control systems use the high volumes of data generated by sensors to identify trends and patterns that can then be used to predict future trends. These predictions help avoid the worst control states making manufacturing processes more efficient while reducing energy consumption. This process enables plants to continuously adapt to new undergoing optimization with no need for operator input. In addition, as the level of inter-connection increases, the AI system can learn from inter-connection, which can lead to the discovery of many complex correlations in systems that aren't yet or are no longer evident to the detection of the traditional control system. Industrial AI control systems with sufficiently intelligent analytical technology are already available. But how to design and develop an AI system depends on the well-studied and thorough understanding of the whole Industrial AI system. This target is motivating us to propose this specific research topic to understand the challenges of Industrial AI and explore how to design and implement an Industrial AI system to optimize industrial control processes.

A traditional industrial control system (ICS) is an information system used to control industrial processes such as manufacturing, product handling, production, and distribution. Industrial control systems include supervisory control and data acquisition systems to control geographically dispersed assets, as well as distributed control systems and embedded control systems to control localized processes. The controller's goal is to move a system from its initial condition to the desired state and, once there, maintain that desired state. The difference between the desired and actual states is called the *error signal*. It is also possible that the desired state will change over time. When this happens, the controller must adjust the state of the system to track changes in the desired state. This human adjustment, of course, works after the error happens, but such afterward adjustment cannot fully optimize the industrial production process on product quality, energy savings, and productivity [19][20].

Built on pattern recognition, Industrial AI proposes a new approach to optimize the industrial control process even further [8]. Interacting with the control system, Industrial AI can predict and avoid potential error. And it also can take over many control functions in the supervisor and controller blocks effectively and efficiently, under different operating conditions. However, human participation plays an important role in the traditional control loop, which can only passively handle any edge case, whereas Industrial AI can intelligently adapt any product line change. Automatic control is always one of the major goals in industrial process applications. In Industrial AI, machine learning enables predictions to take over many control functions in the supervisor and controller blocks. A well-trained machine learning model can provide excellent system performance under widely varying operating conditions. For consistently high performance and robustness, an AI control system must be carefully designed and thoroughly tested [15][34].

# 3  Problems and Petuum Solutions

Industrial AI is being touted as the new electricity that will power the next industrial revolution. Companies that harness AI will lead while those that don't will lag behind. However, the high hurdle of understanding AI, how best to develop and validate AI control system, and ultimately, how to leverage AI as a game-changer create strong barriers to adoption. To ensure industries can truly harness the power of AI for this next revolution, Petuum is transforming the traditional industrial control to Industrial AI.

## 3.1  Problems and Challenges

Although Industrial AI proposes a promising idea for future control systems, manufacturers and solution partners still need to understand how to implement and integrate an AI model into the existing industrial system, thus implementing Industrial AI [13][14]. A well-trained ML model provides many benefits for industrial control optimization; however, a "terrible" Industrial AI design and integration also limits the capability of an ML model. The proposed Industrial AI doesn't give any feasible proposal and details on Industrial AI design and test. Consequently, Industrial AI is being impeded by a bunch of challenges and questions, which need to be clarified to engage more manufacturers embracing this evolution. To better understand how to develop and integrate a trained model into the traditional ICS and test that the deployed AI system is working properly and more effectively than human beings, manufacturers and their AI solution partners need to address all requirements and challenges [23]. Some top challenges and real problems we face when we industrialize AI in manufacturing are discussed below:

1. Domain knowledge transfer. Different manufacturers have various control requirements on their product line. Such requirements include complex domain knowledge, which is difficult for AI solution partners to understand. However, the developed Industrial AI system doesn't work if the trained model cannot fully understand the domain knowledge under the control logic. Therefore, how to translate the industrial domain knowledge to solution partners is the first challenge needed to be addressed.

2. Data preprocessing. Raw data is often incomplete or inconsistent, lacks certain patterns or trends, and likely contains errors. Data preprocessing is the step that transforms raw data into an understandable and expected format in AI training and prediction. In Industrial AI, the data is collected, saved, and processed with streaming mode. The data keeps changing its pattern over time. So, continuously applying the data preprocessing methods in ML training to prediction is the second challenge.

3. AI model accuracy (model validation). AI model accuracy is the measurement used to determine which model is best at identifying relationships and patterns between variables in a dataset based on the input or training data. The better a model can generalize to 'unseen' data, the better predictions and insights it can produce, which in turn deliver more industrial control value.

4. Real-time constraints. In industrial control, real-time constraints are restrictions on the timings of events, such that they occur on-time. The Industrial AI control time includes multiple latencies covering data collection, data transfer, AI prediction, control response, etc. Some of the latencies are constant costs. How to reduce the optimizable latencies to catch end to end real-time constraints is the fourth question needed to be addressed and tested in house.

5. Coexistence of traditional control logic and AI control. Currently, the AI control system is based on its AI model to predict future error. Such prediction can make system avoiding the worst state before it happens, rather than stopping the current worst state. This deviation allows AI control systems to replace traditional control systems gradually. However, machine learning is not a deterministic approach to handle industrial control, which cannot fully satisfy all edge cases in industrial control. Therefore, the traditional control logic still needs to coexist with AI control for extreme situations. This coexistence brings up some new questions, regarding how to define the working scope of AI control and traditional control logic in the industrial system, and whether or not this coexistence brings up decision conflicts in system control.

6. No supervised-steering data for in-house tests. In most cases, the development of new Industrial AI control systems is based on a running product line. The new Industrial AI control system is developed instead of the existing control system. Therefore, there is no any available supervised-steering data or system for in-house testing and debugging. We need to explore the question of how to make in-house testing more feasible for real production line control.

Due to space limitations, only six typical challenges are listed above. All of these challenges need to be addressed in Industrial AI development. In this study, we take the Petuum Optimum system as an example to understand what the challenges for testing and making the AI model productive are and how to address such challenges in an Industrial AI system? And, the generalization potential of this study can benefit the development of Industrial AI systems by helping AI solution partners understand and interpret user requirements better, and improve the procedure of Industrial AI system design, implementation, and test [24].

## 3.2  Petuum Optimum System

Petuum Optimum is a typical industrial control system for industrial manufacturers. The Petuum Optimum system provides a service, which ingests historical and real-time streaming data from the manufacturing plant production line. This data is then leveraged into expert-built machine learning algorithms to deliver precise predictions of the plant's operational variables. The Optimum system conducts precise predictions by understanding the dynamic and non-linear relationships between the variables. Such predictions can prescribe optimal set points to enable the plant to reach its operational goals. Petuum Optimum also offers a supervised-steer option, which enables feeding the set points into the plant's control system for a fully autonomous operation. The self-learning capabilities of Petuum Optimum means it is constantly collecting and processing data to continuously improve the results as it gains a better understanding of the plant and its processes. This paper takes Petuum Optimum as it applies to cement manufacturing plants as an example to discuss the challenges of an Industrial AI. Because performance significantly impacts the system design of an Industrial AI control system, this study mainly focuses on the test of the Industrial AI system. The derived conclusions and experiences can be extended to all of the stages of an Industrial AI system development.

Figure 2 shows an overview of the Petuum Optimum industrial control system and how it integrates with the existing industrial control system. In general, the Petuum Optimum system is serving for prediction and control with three core components - data check, ML model, and control logic. The data check module queries the raw data and the values of the control variable from the controller. The controller caches the raw data collected from sensors and generates dynamic control bounds from the dynamic control table. The data check module enables data preprocessing based on the selected data processing model and sends the preprocessed data the following inference modules based on the raw data and dynamic bounds checking. There are two inference modules in the Petuum Optimum system. If the raw data is checked in bounds in the data check step, ML models are inferred to generate suggestions for industrial control optimization. If the raw data is out of bounds, the control logic module is enabled to take over inference. Control logic acts as a safeguard to guarantee edge case handling. This overview shows the fundamental design and integration logic of the Petuum Optimum system. Petuum Optimum provides a simple system implementation and integration solution for Industrial AI. However, how to validate the developed system works for the plant's optimization goals needs to be carefully and thoroughly tested [21].

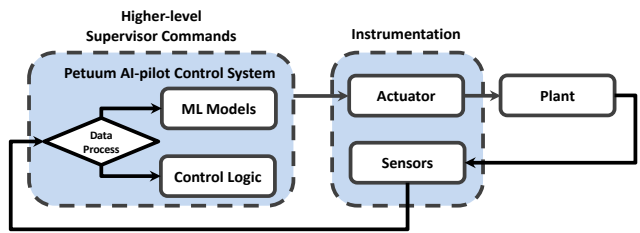

**Figure 2 Petuum Optimum industrial control system**

To better understand how to develop and integrate trained models into the industrial control system, how to test the deployed AI control system, and finally how to outperform traditional systems. Six top challenges include domain knowledge transfer, data validation, AI model performance test, latency constraints, coexistence of traditional control logic and AI control, and in-house test. Petuum Optimum system is used as an example to showcase the challenges in making and testing. Six challenges will be discussed one by one in the following sections.

## 4 Knowledge Transfer

Different manufacturers have various plant physical characteristics and desired outcomes, leading to vastly different control requirements on their product lines. Such requirements include complex domain knowledge that is not clear to solution partners. However, it is impossible to develop an Industrial AI system without adapting it to the customer's specific requirements. Therefore, how to transfer the industrial domain knowledge to AI solution partners need to be addressed first. Knowledge transfer requires excellent and consistent communication between the different parties. In general, it breaks down into three stages.

### 4.1 Symbolize and formulate the knowledge

AI models only accept and process digitalized data during training and inference. In general, the data should be as numeric and tabular as possible to allow easier modeling. With this requirement in mind, the domain knowledge should be formulated and tabulated. The Petuum Optimum system defines an AI Configuration Enablement (ACE) sheet to formalize this step. Three items: selected tags, tag bounds, and tags relation need to be checked and listed in the sheet. These heavily depend on the communication between two parties. It is critical to allow industry partners to transfer their data engineering experience and solution expectation to the AI solution partners.

*Selected tags*

In general, there are hundreds of sensors working on each industrial plant production line. The time series data collected from each sensor is defined as a tag per industry tradition. Data is important, but simply adding more tags does not always lead to a better model. Data engineers often find raw data from tags not qualified for model training because of unsupported format or too much noisy/lost data. Therefore, AI solution partners and end-users should discuss what tags they should keep for AI model training. In general, four types of tags have been defined for every use case in Petuum Optimum system. Control, normalize, and optimize tags can be adjusted to guarantee product line working in optimized status. Constraint tags are used to check whether the product is running in optimized status or an unhealthy way. All of the defined tags are selected based on the four-optimization purpose and target.

*Tag bounds and target values*

Tag bounds define the accepted range of values during optimization. When a tag is outside its predefined bounds or away from its target value, it enters an edge case where 1) the AI model should prioritize returning it to normal, or 2) special safeguard control gets triggered to amend. Tag bounds can be defined as static or dynamic bounds. Four types of tags have been described and discussed in the previous section. The different tag types determine if the tag bounds will be static or dynamic. Control tags can only be changed gradually and are adjusted following dynamic bounds with a small step. Constraint tags are used to check optimization consequences. The static bounds of constraint tags can be defined by customers to check out-of-bounds edge cases. Optimize tags and normalize tags have predefined optimization weight, which suggests different optimization goal.

*Tags relation*

Tags relation describes the positive, negative, and sometimes the non-linear relationship between tags. Such relations are the foundation in traditional Industrial AI. They are the so-called golden rules resulting from physics directly or through years of practice. For example, there are two tags - Tag A and Tag B in the product line. Tag A is a constraint tag, which monitors the product line running status and cannot be adjusted directly. Tag B is a control tag, which can be adjusted to change the production running status. Tag A and Tag B can be defined together to achieve one of the optimize goals. Two tags may have a negative relationship. The relationship can be defined as when the Tag A is increasing quickly, Tag B should be decreased to a low amount to avoid Tag A from continuing to increase.

Understanding how to hand over some of the control logic to the AI model is one of the challenges in Industrial AI. An agreed-upon definition of tag bounds and tag relations can make user expectations clear. The design of the AI model and validation logic among the AI provider party should then fully follow the definition of the ACE sheet.

### 4.2 Disseminate knowledge in multiple teams

In general, An AI solution implementation process involves the collaboration of multiple development teams. Multiple different teams (i.e., customer success, ML, Quality Assurance) need to completely understand user requirements and collaborate coherently to build the Industrial AI solution end to end. For example, the customer success team communicates with the customer, define data labels and expected control logic, and design data preprocessing and basic safeguard control logic. The ML team cares more about how to tune the selected model to catch data features with fully requirements understanding. And, the QA team also needs to completely understand the user requirements and design different test cases to validate the solution from the customer's perspective. Therefore, all of the involved teams should keep customer expectations and transferred domain knowledge consistent. Once the product requirement document (PRD) and ACE have been defined, any update should be re-checked with customers.

### 4.3 Re-check knowledge in supervised-steering test

In our experience, even with a well-defined ACE, all of the knowledge and expectations are not necessarily transferred. More often than not, customers may not fully describe their knowledge and expectations in one iteration, and the AI solution partner does not fully understand the user description either. Many intrinsic reasons could lead to such suboptimal knowledge transfer, and it is in the best interest of both sides to continue communication during and after the ML modeling. Therefore, it is advised to re-check knowledge and expectations with the user in a supervised-steering test. In this phase, the drafted Industrial AI solution is shown to the users and is a chance for them to provide second-round feedback. The re-checking helps the user and solution partner correct the ACE and PRD and fine-tune the deployed solution.

## 5 Data Validation

Machine learning is a powerful learning tool that extracts patterns from data. By this nature, a machine learning model is acutely sensitive to the quality of the data. Such sensitivity is relevant in the two core phases of ML model development: training and deployment [2][3].

During training, data quality directly determines the ceiling of model performance. After all, ML models, no matter how powerful, cannot learn a pattern if it is not present in the data. A common challenge ML engineers face is that they do not observe the data patterns mentioned by the customers. While these usually involve further discussion and debugging, there are several common data preprocessing steps that are general enough to be applied in almost all use cases. Below, we introduce the four common data preprocessing topics in the Petuum Optimum system.

### 5.1 Outlier removal

Outliers are extreme values that deviate from other observations on data. In some cases, customers provide clear definitions for outliers, such as a threshold that defines the normal range of values. In most cases, however, this information is not readily available, and data engineering is needed [10]. Descriptive statistics methods, such as the interdecile range, checks data distribution between the first and the ninth deciles (10% and 90%). Outliers are padded with the last normal value. If outliers exist too often, data should be re-verified with customers before proceeding to ML training. It catches potential errors early, allowing fast iteration of communication between customers and solution providers. Sometimes, such re-verification also helps ML engineers improve their outlier removal method.

### 5.2 Non-stationary signals

ML models are trained on a limited amount of data. The data pattern is assumed to be persistent in going outside the training dataset. Unfortunately, this is sometimes not the case. The data pattern needs to be continuously monitored and validated throughout the various stages of the Industrial AI system. Otherwise, model performance may deteriorate over time. When overlooked, it may even lead to an unexpected abrupt drop in performance.

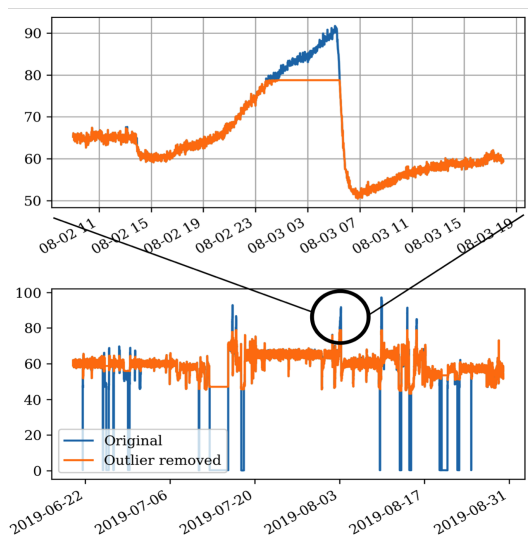

**Figure 3 Outlier removal example**

Figure 3 showcases a real example that occurred during Petuum Optimum development. The original signal (blue) was outlier removed (red), as shown in the top panel. Upon closer inspection, however, we realized that the 80s values on August 3 are well within the expected range of this signal, therefore, should not be removed. Further debugging revealed that they were labeled as outliers because the model was trained based on the data of June. As shown in the bottom panel, data in June mostly fluctuated around 60 and never exceeded 70. Because of this shift in data pattern, models performed very poorly for hours in August. Fortunately, this was captured in the validation stage before deployment, which underlines the importance of continuous monitoring of data quality in case the signal is non-stationary.

## 5.3 Broken sensors

In the previous section, we discussed outliers. In some production lines, the sensors may generate outlying values often enough that they are difficult to be accurately identified by statistics. These sensors may restart periodically to clean or re-calibrate. A dedicated data engineering team is required to work with broken sensors. This situation is formulated as follows:

At each step, a window of data is injected into the ML model for inference:

$$X = [x_{t-w+1}, x_{t-w+2}, \ldots, x_t]$$

Where $x_t$ represents the values of a signal at time t, and w is the window size. Petuum Optimum uses a majority vote to decide whether a window is valid. In a given window, if the number of data points demonstrating broken characteristics is more than 25%, this window is considered unstable, and no further ML effort will be made on this window.

To determine whether the sensor is working in one specific data point, customers shall provide a criterion, for example, a "broken sensor bound." If the sensor value is within the bounds, it is considered valid. Let's take one tag 'Tag Name 1' as an example to extend the details. Take a specific window of [5.0, 1.0, 4.0, 7.5, 4.5, 3.8, 16.2, 17.2, 36.0, 44.0, 2.0, 77.0] for example. Here, the window size is 12, and the broken sensor bound is [15, 50]. We observe that eight data points are out of the broken sensor bounds. Consequently, 67% (8/12) of the members in this window are invalid, which is above the 25% limit making the window unstable.

## 5.4 Interpolation

Streaming is the most common data source in industrial control. Streaming data is generated continuously by many sensors, which typically are sent in small sizes simultaneously. It is common for different sensors to generate data with different sampling rates. Different data streams are interpolated to the same sampling rate to simplify the data processing pipeline. The same interpolation is used in both ML model training and inference [6].

If the customer suggests the interpolation method, it can be implemented directly. Otherwise, Petuum Optimum defaults to a simple interpolation method that always writes the most recently cached value to the current timestamp. This method is extremely easy to implement, executes fast, and works well with stringent time requirements. Figure 4 shows two examples of interpolation.

Data processing is an integral step in ML. ML engineers may not have the domain knowledge to comprehend the data, whereas customers may not understand what is needed for successful ML training and deployment. Therefore, the data processing method and output should be discussed and verified between the two parties frequently [7].

## 6 Machine Learning Model Performance Test

From quality and testing perspective, developing Industrial AI system is no difference from traditional software development, in the sense that it is also a full life cycle involving unit test, integration test, and end-to-end test [9][12]. However, there are three major differences between software development and ML development which makes testing approach slightly different. 1) ML model is not a deterministic approach, and it cannot generate the same output for every test run hence defining pass/fail criteria is challenging. 2) the performance of ML model heavily depends on the data quality in training and serving. ML model should be able to catch data pattern changes with parameters updates automatically. 3) Industrial AI is developed and deployed to serve for streaming application. The ML model performance could be significantly impact by streaming latency. We need a new testing approach to check and validate the impact of such latency [17].

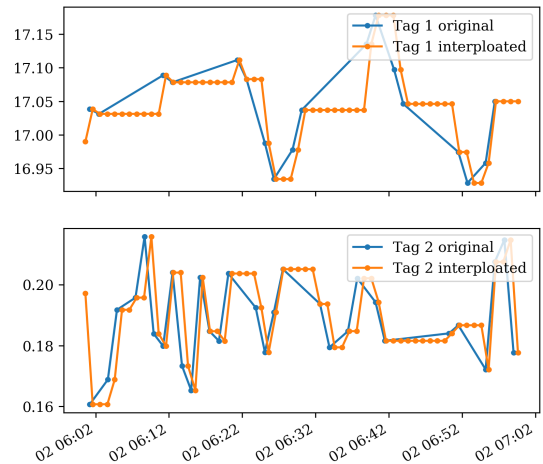

**Figure 4 Interpolation example**

The goal of the Industrial AI system is to prescribe the optimized value to the controllers in production. This task is divided into two components: 1) a prediction model that forecasts production line behavior in the future, usually after a certain number of minutes; 2) an optimization model which finds the best control value to the controller and optimizes the future behavior. When framed as an ML task, there are two types of variables in the model: response variables and control variables. The response variables, which are measured by sensors, represent the current status of the system under control. The control variables are control components set up by the actuator to make the system reach the desired status. The two types of data have a casual relation due to their physical origin, which should be learned by the prediction model [29][31]. To optimize response variables to the desired status, Petuum Optimum searches for the best control variable with trained prediction model evaluation and feeds the control variables into the system to optimize response variables. Concretely, the ML task is formulated as follows:

$X_t$: the response values of input values at instant t;

$Y_t$: the control values of input values at instant t;

$Z_t$: the values of output values at instant t;

In the regression model, the formula:

$\vec{X}_t = [X_{t-w+1}, X_{t-w+2}, \ldots, X_t]$ and $\vec{Y}_t = [Y_{t-w+1}, Y_{t-w+2}, \ldots, Y_t]$ are used to predict $Z_{t+\Delta t}$, where w is the window size, $\Delta t$ is the prediction length. Let $\hat{Z}_{t+\Delta t} = F(\vec{X}_t, \vec{Y}_t)$, then the optimize target is

$$\min \left(\sum_t (\hat{Z}_{t+\Delta t} - Z_{t+\Delta t})^2\right), \text{ i.e., let } F(\vec{X}_t, \vec{Y}_t)$$

as close to the actual output $Z_{t+\Delta t}$ as possible. Consequently, the loss function of the regression algorithm is defined as:

$$\text{loss function} = \sum_t (\hat{Z}_{t+\Delta t} - Z_{t+\Delta t})^2$$

Two test cases are used to validate the model performance. First, an offline evaluation is conducted to verify the performance of the trained model. Once a model is trained with some training data, the test chooses historical data from a different period with the same preprocessing process. The prediction results are compared with the raw data to validate the prediction model's accuracy (Figure 5). The prediction signal is shifted by the prediction length to match the actual data. The example shows a bad performance for the prediction model. The prediction often does not catch extreme values in the original data. Through visualization, together with other numerical criteria such as the correlation, r squared, and mean square error (MSE), we can confidently reject this model and investigate ways to improve its performance.

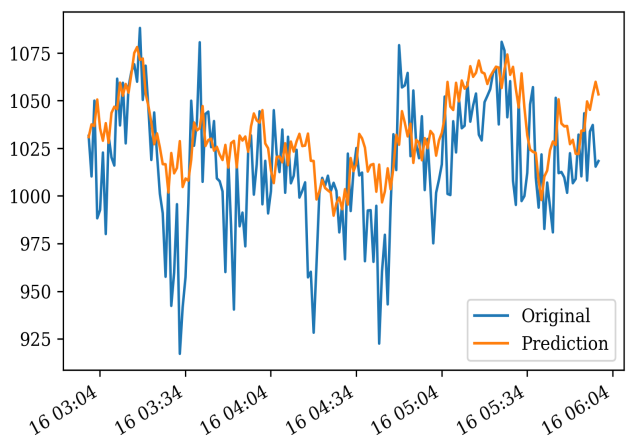

**Figure 5 Prediction results offline evaluation**

The second test case is an online streaming evaluation, which completes the test by checking performance in a semi-real deployment. Many factors impact the performance of the deployed model. First, the data pattern could change. Additionally, the steaming infrastructure could impact the data quality because of latency, missing samples, etc. Specifically, the latency of each sensor could be different, and they may significantly impact the model behavior.

## 7 Latency

Time is of the essence in industrial control automation. Real-time control requires low latency throughout the pipeline, including potential latency introduced during data collection, data transfer, ML prediction and control response. Some of the latencies are fixed costs, such as data collection or transfer. Two typical latencies – data subscription and ML inference are optimized in the Petuum Optimum system [28].

Latency occurs in every step during data streaming, which is based on a publish-subscribe structure. In the publish-subscribe system, the customer's plants collect and write data into a data warehouse [35][36]. A publisher then reads the time series data from the data warehouse and produces messages with specified topics. Every AI solution needs to subscribe to their required data with a defined topic from the publisher. In the publish-subscribe system, many factors, such as topic numbers, system cache, and subscriber numbers, could potentially affect the end-to-end latency and thus change the behavior of the ML model. Therefore, the deployed data pipeline needs to be fully designed and tested with strict time constraints [32][33].

Another major source of latency is the optimization model. At each time instance, the best control values are prescribed by the ML model. The optimization process finds the best value among the solution space but is usually very time-consuming, so shortening the time complexity is pivotal in this step. For example, reducing the inference time from one minute to eight seconds for each iteration is possible with parallel computing. The inference speed is not strictly part of the core performance of ML models, but it is a key metric to evaluate how practical the product is. Customers may also define their requirements in the PRD.

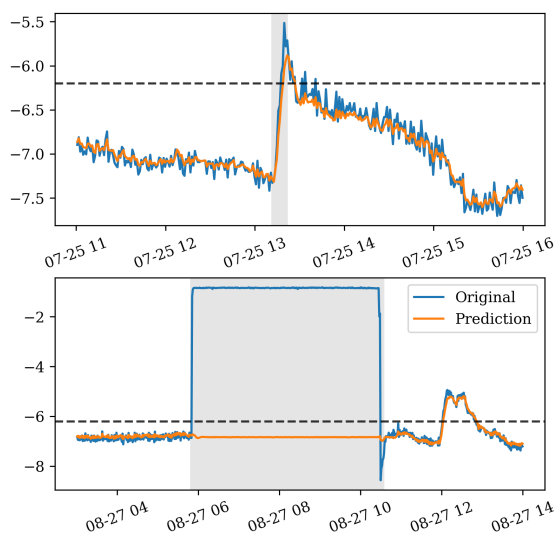

**Figure 6 The streaming data with system latency**

Figure 6 is an example of a tag with occasional sharp spikes (top panel) within minutes. Because of network errors, prolonged data loss occurred for hours at one point, which can dramatically affect model performance if not resolved.

## 8 Coexistence of traditional control logic and AI control

The Industrial AI control system uses its ML model to predict the future state and avoid undesirable outcomes. Such prediction is key as it preempts an undesirable state before it happens rather than mitigating afterward. This distinction gives AI control systems a tremendous advantage over traditional control systems. However, ML is often not a deterministic approach, and mistakes may occur. More importantly, the very nature of edge cases means they have little or no representation in the training data. In the current stage, it is unwise to give ML models full control of the plant. We advise the traditional control logic to coexist with AI for unforeseen corner cases. This coexistence brings up new questions as to how the scopes of AI and traditional control logic fit in the industrial system. It is also important to resolve conflicts between the two when differences arise.

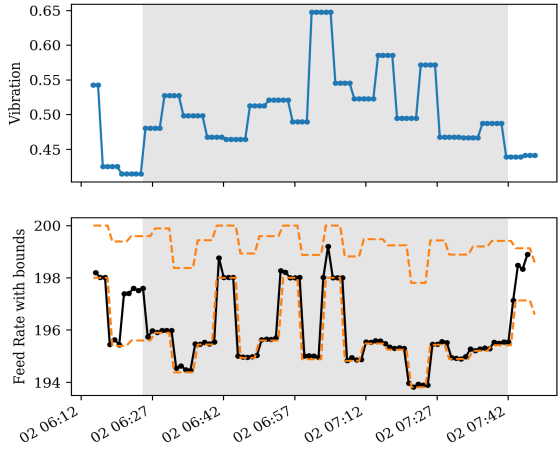

**Figure 7 Control logic validation**

We use the industrial control for the vertical mill, a typical grinder used in cement manufacturing, to illustrate the challenge and our solution of AI coexistence. The mill's body vibration is one of the main metrics necessary to monitor for safe production. Several reasons lead to high vibrations, such as failure of the grout, distortion of the steel structure, or cold joints between pours [1]. The ML model cannot fully predict all of these factors and their interaction because of the lack of data and long reaction time. To ensure safety, safeguard logics based on expert knowledge in traditional industrial control logic are still in use. In the Petuum Optimum system, the model enters the survival mode if the current vibration is over a defined threshold of 0.45 (Figure 7 Top). The AI model is required to perform everything within its power to reduce vibration. Figure 7 Bottom shows how special safeguard logic takes over in survival mode. The feed rate prescription is mostly maintained at the lowest level possible allowed by the system (indicated by dash bounds) because domain knowledge from the customer tells us low feed rate alleviates the vibration.

## 9   In-house Test

Traditionally, the improvement of an industrial control system is incremental. The Petuum industrial Optimum system is, however, developed from the ground up and not on the running product line directly. This poses a big restriction on the model validation during development: as the AI solution partner, we do not have direct access to the controllers in the plant for both business and safety reasons. How to evaluate a trained ML model before its release and deployment is, therefore, a critical challenge for Industrial AI testing. We adopted a one-step verification to validate model output (prediction and optimization) at each timestamp. Instead of looking for cumulative effects over time, we examine whether the outputs are optimized in the correct direction during each step. This is because the AI model suggestion cannot apply to the actual input of the system without physically controlling the production line. An in-house test is necessary to thoroughly test the speed and amplitude of the optimization over many steps. During in-house testing, AI models are first deployed with close supervision from both model developers, quality engineers, and industry experts internally. One-step verification is not a replacement for in-house verification but a convenient testing framework that helps us validate the model's offline behavior.

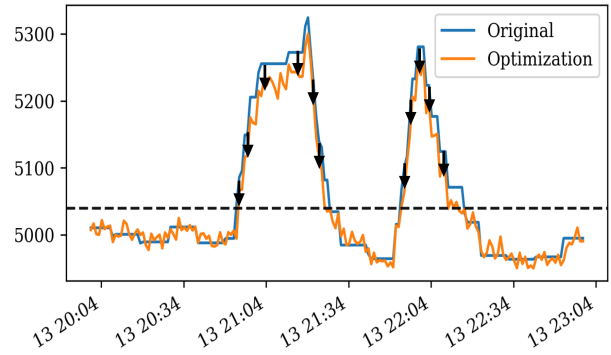

**Figure 8 One-step validation**

Figure 8 demonstrates an example of one-step validation. The shown signal has an upper constraint bound of 5050, meaning it should be decreased when above this threshold. Upon validation, we observe that optimization towards the lower side does occur whenever this signal is above constraint. If this were in-house testing with an actual deployment, we would expect the original measurement to gradually decrease until back within normal range. This is, of course, not possible in offline testing, but one step validation offers a basic level of confidence in the model's optimization power. Due to the limitation with in-house testing, to mitigate the risk before we put the model in auto-steer model in production, we use on-site testing approach and put model in non auto-steering mode in production. This way we can monitor how the model perform with multiple step change and side-by-side compare the results with traditional control system. And because we have done thorough in-house testing to validate the models, this final step only takes very short amount of time before we switch to auto-steer mode in production.

## 10   Conclusion and Future Works

The concept of Industrial AI is becoming widely adopted in mainstream manufacturers in their discourse and practice. A well-studied industrial system can provide excellent system performance under widely varying operating conditions. Machine learning enabled predictions and optimization hold a strong advantage over traditional control methods. To ensure a consistently high level of performance and robustness, an AI control system must be carefully designed and thoroughly tested. In the study, we present the Petuum Optimum system as an example to demonstrate the challenges in Industrial AI deployment, and more importantly, how to solve and validate them. Six challenges from our real experience when implementing Industrial AI were discussed. The challenges showcase the implementation details of Industrial AI. Based on the findings of challenges, four takeaways were delivered to summarize the study.

1. Data quality directly determines the performance of AI model. Data labeling, preprocessing, and quality validation should be checked and designed carefully;
2. The industrial product line is a dynamic system. The running behavior and patterns are always changing. The trained model should catch such changes to better serve live deployment;
3. The implementation of Industrial AI includes the trained model itself and a deployed running system. The model performance

could be changed in the running system. A well-designed test should cover both situations;

4. Thorough testing and various testing approaches are required to ensure the high quality of the model and to detect corner cases before the model is deployed to the production environment. Any issues missed in the testing process may cause sensor hardware damage and tremendous loss to the factory as well as the AI company.

A major limitation of this study is its exclusiveness of the type of data. Petuum Industrial AI is designed for typical industrial control in manufacturer plants that involve multi-channel time series data. Multimedia materials such as images and text will be interesting topics worth further exploration. In addition, some steps, such as knowledge transfer, involve some manual configurations and updates. Such manual configurations are error-prone [16].

The limitations will be removed gradually. Our future work includes 1) extending our Industrial AI solution to a wide variety of time series data types, such as speech and video. Such time series data have different data processing policies, latency requirements, and model selection. 2) removing manual configurations in solution development and validation. Automation code can be added to create configuration files based on predefined templates. Such improvement can significantly reduce the time and cost spent on development and validation.

We summarized our challenges and experience in developing Industrial AI in this study. Ultimately, it shall serve as a foundation for the wide adoption of AI technology in the field of industrial control.